\documentclass[10pt, a4paper]{article}

\usepackage[final]{lrec2026} 

\usepackage{times}
\usepackage{latexsym}
\usepackage[T1]{fontenc}
\usepackage[utf8]{inputenc}
\usepackage{microtype}
\usepackage{inconsolata}
\usepackage{graphicx}
\usepackage{amsmath}
\usepackage{amsfonts}
\usepackage{booktabs}
\usepackage{algorithm}
\usepackage{algorithmic}
\usepackage{multirow}
\usepackage{url}
\usepackage{subcaption}

\title{Transformer-Enabled Diachronic Analysis of Vedic Sanskrit: \\
Neural Methods for Quantifying Types of Language Change}

\name{Ananth Hariharan, David Mortensen} 

\address{University of Illinois Urbana-Champaign, Carnegie Mellon University \\
         ahari8@illinois.edu, dmortens@cs.cmu.edu}

\abstract{
This study demonstrates how hybrid neural-symbolic methods can yield significant new insights into the evolution of a morphologically rich, low-resource language. We challenge the na\"ive assumption that linguistic change is simplification by quantitatively analyzing over 2,000 years of Sanskrit, demonstrating how weakly-supervised hybrid methods can yield new insights into the evolution of morphologically rich, low-resource languages. Our approach addresses data scarcity through weak supervision, using 100+ high-precision regex patterns to generate pseudo-labels for fine-tuning a multilingual BERT. We then fuse symbolic and neural outputs via a novel confidence-weighted ensemble, creating a system that is both scalable and interpretable. Applying this framework to a 1.47-million-word diachronic corpus, our ensemble achieves a 52.4\% overall feature detection rate. Our findings reveal that Sanskrit's overall morphological complexity does not decrease but is instead dynamically redistributed: while earlier verbal features show cyclical patterns of decline, complexity shifts to other domains, evidenced by a dramatic expansion in compounding and the emergence of new philosophical terminology. Critically, our system produces well-calibrated uncertainty estimates, with confidence strongly correlating with accuracy (Pearson r = 0.92) and low overall calibration error (ECE = 0.043), bolstering the reliability of these findings for computational philology.
 \\ \newline }

\begin{document}

\maketitleabstract

\section{Introduction}

Many older people, of every generation, complain that the speech of younger people has become less complex and less expressive, reflecting an ideology in which language change is simplification. Philologists---scholars who study old texts---have often been subject to the same prejudice. They have seen changes in the complexity of morphology, from Latin to Romance languages, from Ancient Greek to modern Greek, and from Sanskrit to modern Indo-Aryan languages, through the lens of global simplification.

Do languages always get simpler over time?
Vedic Sanskrit offers an ideal laboratory to test this hypothesis. As one of the oldest attested Indo-European languages, its 2,000+ year evolution from the poetic hymns of the Rigveda (c. 1500 BCE) to the philosophical prose of the Upanishads and Classical texts (c. 500 CE) presents a rich tapestry of linguistic change. However, analyzing this evolution computationally poses immense challenges. Like many historical languages, Sanskrit is low-resource, lacking the large, annotated datasets that fuel modern NLP. Furthermore, its rich fusional morphology and extensive use of euphonic combination (sandhi) make even basic tasks like tokenization non-trivial, preventing standard models from being applied effectively.

To overcome these barriers, we introduce a novel, weakly-supervised framework designed specifically for diachronic analysis in morphologically rich, low-resource settings. Our hybrid approach first leverages a small set of high-precision linguistic patterns encoded as regular expressions to automatically generate thousands of ``pseudo-labeled'' examples from unannotated text. We then use this synthetic data to fine-tune a transformer model, creating a robust feature detector capable of recognizing subtle contextual patterns. By ensembling this neural model with our original symbolic rules, we create a system that is both scalable and relatively interpretable.

Applying our novel weakly-supervised hybrid framework to a diachronic corpus of 20 Sanskrit texts spanning 1500 BCE to 500 CE, we find that while specific features ebb and flow, overall complexity is often redistributed rather than simply lost (consistent with the findings of \citet{shim-etal-2024-phonotactic} for Dutch phonology). This finding reinforces the conclusion of \citet{pimentel-etal-2020-phonotactic} that there is a trade-off in complexity among some aspects of grammar. Internal validation shows substantial agreement between these components, and critically, our system produces reliable uncertainty estimates, addressing a key need for interpretability in computational philology.

Beyond Sanskrit, the framework shows potential for adaptation to other historical languages that share similar typological and data constraints. It is particularly well-suited for languages with rich inflectional morphology, such as Latin, Ancient Greek, and Old Church Slavonic; for those with limited annotated corpora, including Hittite, Avestan, and Gothic; and for languages with well-documented diachronic stages, such as Arabic, Chinese, and Tamil.

Our primary contributions thus include: (1) a validated neural-symbolic method with calibrated confidence weighting for analyzing low-resource historical languages; (2) quantitative insights into Sanskrit's morphological evolution, specifically the redistribution of complexity; and (3) the release of our curated diachronic corpus, trained models, and feature extraction pipeline to facilitate reproducible research in AI-augmented historical linguistics.

\section{Related Work}
\label{sec:related-work}

Our research is situated at the intersection of computational historical linguistics, low-resource NLP, and the computational analysis of Sanskrit. We build upon previous work in each of these domains.

\subsection{Computational Historical Linguistics}

Early computational approaches to historical linguistics primarily employed statistical methods \citep{bouchard2013automated} or phylogenetic algorithms \citep{gray2003language} to model language evolution. More recently, the field has seen a significant shift toward applying transformer-based models to classic linguistic problems. Researchers have successfully used transformers for tasks like proto-language reconstruction, demonstrating their effectiveness even in data-scarce scenarios \citep{kim2023transformed, lu-etal-2024-improved}.  This trend extends to modeling the complex reasoning involved in historical linguistic analysis, such as framing multi-step sound law induction as a Programming by Examples (PBE) task, which remains challenging even for state-of-the-art reasoning models \citep{naik2025programming}.

Ensuring large language models respect temporal boundaries without ``knowledge leakage'' is crucial for diachronic analysis. While pretraining separate period-specific models offers one solution \citep{fittschen2025pretraining}, our work employs a single, unified framework. This model learns to distinguish historical periods via a multi-task objective, focusing on large-scale feature detection to quantify linguistic change and complementing other transformer-based efforts.

\subsection{NLP for Low-Resource Languages}

A key challenge in applying modern neural methods to historical languages is data scarcity. For low-resource scenarios, researchers have demonstrated the effectiveness of transfer learning \citep{kann2017one} and hybrid approaches that combine neural and symbolic methods for tasks like morphological inflection \citep{mccarthy2019marrying}. Our work addresses the data scarcity problem through a different lens: we use a weakly-supervised, hybrid neural-symbolic approach to generate training data and build a robust feature detection system from a large, unannotated corpus.

\subsection{Computational Analysis of Sanskrit}

The application of NLP to Sanskrit presents unique challenges due to its rich fusional morphology, relatively free word order, and significant linguistic variation across its long history. Even foundational tasks like word segmentation are non-trivial due to pervasive \textit{sandhi} (euphonic combination), which has motivated the development of specialized, transformer-based tokenizers to achieve the current state-of-the-art in performance \citep{sandhan-etal-2022-translist}. 
This complexity necessitates a focus on creating robust resources and applying sophisticated data-driven methods to overcome these hurdles.

Foundational work in our specific domain of \textbf{Vedic Sanskrit} has seen the creation of several critical resources. These include a metrically restored and phonologically annotated digital text of the \textit{Rig Veda} \citep{VanNooten1995RigVeda}, the Vedic Treebank (VTB) for syntactic analysis \citep{hellwig-etal-2020-treebank}, and the {Vedavani} corpus, a 54-hour dataset for Vedic poetry ASR \citep{kumar2025vedavani}. Building on the VTB, \citet{hellwig-etal-2023-data} presented the first data-driven dependency parser for Vedic, finding that in a low-resource setting, high-quality morphosyntactic features are crucial for performance. Our work contributes a large-scale diachronic text corpus --- complementing these syntactic, phonological, and audio resources --- and emphasizes morphological feature detection rather than full syntactic parsing.

Computational approaches to Sanskrit have further tackled its data scarcity and morphological richness through various strategies. Comprehensive evaluations confirm that contextual embedding models like BERT effectively represent Sanskrit, justifying their use in downstream applications \citep{sandhan-etal-2023-evaluating}. Unsupervised methods have also been explored; for instance, \citet{srivastava-etal-2018-deep} demonstrated the utility of a BiLSTM autoencoder for POS tagging, underscoring the importance of character n-grams for capturing the language's complex morphology. 

Research has also yielded systems for more specialized tasks, such as efficient prosodic analysis \citep{shastry-wali-2023-machine}. Notably, even classical NLP methods like LSA have sometimes outperformed modern deep learning embeddings for specific tasks like topic modeling the Rigveda \citep{bollineni2025mapping}. Our work builds upon these collective insights, particularly leveraging the demonstrated representational power of transformers, but applies them within a novel weakly-supervised, hybrid model specifically designed to quantify historical linguistic change at scale.

\section{Methodology}

\subsection{Data \& Feature Set}

Our corpus comprises 20 texts totaling 1,474,356 words across four chronological periods, shown in Table \ref{tab:corpus}. The Samhitas correlate with the Early Vedic period, the Brahmanas with the Late Vedic period, and the Upanishads with the Latest Vedic period.

\begin{table}[h]
\centering
\small
\begin{tabular}{lrr}
\toprule
\textbf{Period} & \textbf{Texts} & \textbf{Words} \\
\midrule
Samhitas (1500-1000 BCE) & 6 & 590,283 \\
Brahmanas (1000-700 BCE) & 5 & 321,284 \\
Upanishads (700-300 BCE) & 6 & 36,192 \\
Classical (300 BCE-500 CE) & 3 & 526,597 \\
\bottomrule
\end{tabular}
\caption{Corpus composition by historical period}
\label{tab:corpus}
\end{table}

Texts are transliterated using the International Alphabet of Sanskrit Transliteration (IAST) with UTF-8 encoding and undergo lowercase normalization. The raw texts for our 20-text corpus  were sourced from the Göttingen Register of Electronic Texts in Indian Languages (GRETIL). Our hybrid framework employs two distinct tokenization strategies tailored to its components. For the symbolic analysis, we use Unicode-aware word-level tokenization so that regex patterns can be applied to complete words. For the neural component, the text is processed by mBERT's native WordPiece tokenizer, which segments words into subword units. This subword tokenization is crucial, as it allows the transformer to effectively model Sanskrit's complex morphology and handle out-of-vocabulary words.

We analyze 163 linguistic features across five domains:

\begin{itemize}
\item \textbf{Phonological Evolution (15 features)}: Phonological Evolution (15 features): Early Vedic phonology (retroflex ḷ; pluti vowels – overlong trimoraic vowels); vowel system changes (diphthongs ai/au → monophthongs e/o); consonant patterns (retroflex assimilation, cluster simplification).

\item \textbf{Morphological Systems (78 features)}: Recessive verbal systems (subjunctive paradigms, injunctive forms, reduplicated presents), nominal evolution (dual decline, case system changes), innovative constructions (periphrastic perfects, modal formations)

  \item \textbf{Syntactic Development (35 features)}: Particle systems (Vedic discourse markers \textit{sma}, \textit{ha}, \textit{vai}), participial
  constructions, correlative patterns, subordination complexity
  \item \textbf{Lexical Stratification (25 features)}: Religious terminology evolution, philosophical vocabulary emergence,
  substrate borrowings, compound elaboration
  \item \textbf{Stylistic Markers (10 features)}: Genre indicators, discourse patterns, metrical features
\end{itemize}

Each linguistic feature pattern is implemented as a Unicode-aware regular expression ($r_i$) combined with patterns that check the surrounding text context. Specifically, we define positive context patterns ($C^+_i$) that must appear nearby and negative context patterns ($C^-_i$) that must not appear nearby to increase the confidence of a match. The overall pattern structure can be represented as:

\begin{equation}
P_i = \langle r_i, C^+_i, C^-_i, w_i \rangle
\end{equation}

Here, $r_i$ is the base regex matching the target feature itself. $C^+_i$ and $C^-_i$ are sets of regular expressions representing the positive and negative contextual cues, respectively. These context patterns are checked within a $\pm$20-word window surrounding a potential match of $r_i$. The final component, $w_i$, is a confidence weight calculated based on the context: $w_i = \text{clamp}_{[0.1,0.95]}(0.6 + 0.2|C^+_i|_{\text{matched}} - 0.3|C^-_i|_{\text{matched}})$, where $|C^+_i|_{\text{matched}}$ and $|C^-_i|_{\text{matched}}$ are the counts of matched positive and negative context patterns within the window. A minimum confidence threshold of 0.4 is applied for a pattern match to be included in the weak supervision data.

\subsection{Transformer Architecture}

We employ \textbf{multilingual BERT (mBERT)} \cite{devlin2019bert} as the base encoder, leveraging its cross-lingual representations and subword tokenization to handle the morphological complexity of Sanskrit. Given
an input token sequence
\[
\mathbf{X} = [x_1, x_2, \ldots, x_L],
\]
where $L$ is the sequence length and each $x_i$ is a WordPiece token, the BERT encoder produces a sequence of contextualized hidden representations:
\[
\mathbf{H} = \text{BERT}(\mathbf{X}; \theta_{\text{BERT}}),
\]
where $\theta_{\text{BERT}}$ denotes the parameters of the pretrained BERT model and
$\mathbf{H} = [\mathbf{h}_1, \mathbf{h}_2, \ldots, \mathbf{h}_L] \in \mathbb{R}^{L \times d}$,
with $d = 768$ representing the hidden dimension.
The hidden state corresponding to the special [CLS] token, denoted $\mathbf{h}_{\text{CLS}} \in \mathbb{R}^d$, serves as a holistic sequence representation for downstream tasks.

We attach two task-specific heads to $\mathbf{h}_{\text{CLS}}$, each predicting distinct outputs through independent parameter sets.

A two-layer feed-forward network with dropout predicts independent probabilities for each of $K_m$ morphological features:
\[
\mathbf{z} = \mathbf{W}_2 \cdot \text{ReLU}(\mathbf{W}_1 \cdot \mathbf{h}_{\text{CLS}} + \mathbf{b}_1) + \mathbf{b}_2,
\]
\[
\mathbf{p}_{\text{morph}} = \sigma\left(\frac{\mathbf{z}}{\tau}\right),
\]
where $\mathbf{W}_1 \in \mathbb{R}^{(d/2) \times d}$, $\mathbf{W}_2 \in \mathbb{R}^{K_m \times (d/2)}$, $\sigma(\cdot)$ is the element-wise sigmoid, and $\tau$ is a learnable temperature parameter for calibration.
Dropout with rate 0.1 is applied after the ReLU activation. We use binary cross-entropy loss (PyTorch \texttt{BCEWithLogitsLoss}) to enable multi-label prediction, allowing texts to exhibit multiple morphological features
simultaneously. Subsequently, a two-layer network estimates prediction confidence:
\[
c = \sigma(\mathbf{w}_2^\top \cdot \text{ReLU}(\mathbf{W}_1^c \cdot \mathbf{h}_{\text{CLS}} + \mathbf{b}_1^c) + b_2),
\]
where $\mathbf{W}_1^c \in \mathbb{R}^{(d/4) \times d}$, $\mathbf{w}_2 \in \mathbb{R}^{d/4}$, and $\sigma(\cdot)$ produces $c \in [0, 1]$, with dropout rate 0.1.

\subsection{Weak Supervision Training}

We automatically generate training labels using our regex patterns with confidence adjustment:

\begin{algorithm}
\caption{Weak Supervision Label Generation}
\begin{algorithmic}[1]
\FOR{each word $w$ in corpus}
\STATE Extract context window $C_w$
\FOR{each pattern $P_i$}
\IF{$r_i$ matches $w$}
\STATE $c_{base} \leftarrow 0.6$
\FOR{each $p \in C^+_i$}
\IF{$p$ matches $C_w$}
\STATE $c_{base} \leftarrow c_{base} + 0.2$
\ENDIF
\ENDFOR
\FOR{each $n \in C^-_i$}
\IF{$n$ matches $C_w$}
\STATE $c_{base} \leftarrow c_{base} - 0.3$
\ENDIF
\ENDFOR
\STATE $c_{final} \leftarrow \min(0.95, \max(0.1, c_{base}))$
\STATE Add label $(w, C_w, P_i, c_{final})$
\ENDIF
\ENDFOR
\ENDFOR
\end{algorithmic}
\end{algorithm}

This process generates approximately 50,000 word-context pairs with morphological labels and confidence scores.

\subsection{Confidence-Weighted Ensemble}

We combine transformer predictions with regex patterns using confidence weighting:

\begin{equation}
f_{\text{ensemble}} = \frac{w_t \cdot c \cdot f_t + w_r \cdot f_r}{w_t \cdot c + w_r}
\end{equation}

where $f_t$ and $f_r$ are transformer and regex frequencies, $c$ is the transformer confidence, and $w_t$, $w_r$ are method-specific weights determined empirically for each feature category. We verified that the weighted combination preserves normalized frequency distributions (sum to 1) via renormalization.

\subsection{Evaluation Framework}

To evaluate our framework, we employ multiple complementary validation approaches. Our primary evaluation uses pattern detection agreement analysis, comparing predictions between the regex-based symbolic system and the fine-tuned transformer across all 20 texts in our corpus. Agreement is defined as instances where both methods detect the same linguistic feature, with correlation computed using Pearson's $r$. We also construct a linguistic gold standard validation set based on established Sanskrit grammatical scholarship (Whitney's Sanskrit Grammar \citep{whitney1889sanskrit} and Macdonell's Vedic Grammar \citep{macdonell1916vedic}), consisting of manually annotated morphological analyses for key features including subjunctive verb forms, reduplicated perfects, dual case endings, emphatic particles, and philosophical vocabulary. Each gold standard example includes the target word, sentential context, true morphological analysis with confidence scores, expected false positives, and distinguishing contextual features. For confidence calibration, we measure the correlation between the model's predicted confidence scores and empirical classification accuracy using Pearson correlation and the Expected Calibration Error (ECE).

Our task formulation is as follows: Given a Sanskrit text, the system must detect the presence and frequency of the specified morphological, syntactic, and lexical features normalized per 1,000 words and assign a confidence score to each prediction. Pattern detection agreement is measured as:

$$\text{Agreement} = \begin{cases}
1 & \text{if } \frac{|f_{\text{regex}} - f_{\text{transformer}}|}{\max(f_{\text{regex}}, f_{\text{transformer}})}
< 0.30\\
0 & \text{otherwise}
\end{cases}$$

where $f_{\text{regex}}$ and $f_{\text{transformer}}$ are the normalized feature frequencies (per 1,000 words) detected by each method. Confidence calibration is quantified via:

$$r_{\text{Pearson}} = \text{corr}(\mathbf{c}{\text{pred}}, \mathbf{a}{\text{true}})$$

where $\mathbf{c}{\text{pred}}$ is the vector of predicted confidence scores and $\mathbf{a}{\text{true}}$ is the vector of empirical accuracies. Expected Calibration Error is computed by partitioning predictions into $B$ bins and calculating:

$$\text{ECE} = \sum_{b=1}^{B} \frac{n_b}{N} \left| \text{acc}(b) - \text{conf}(b) \right|$$

where $n_b$ is the number of predictions in bin $b$, $N$ is the total number of predictions, $\text{acc}(b)$ is
the empirical accuracy within bin $b$, and $\text{conf}(b)$ is the average predicted confidence in bin $b$.

\subsection{Statistical Analysis}

Diachronic trends are quantified using multiple complementary approaches to ensure robust statistical inference. We employ ordinary least squares linear regression where texts are ordered chronologically and regression coefficients indicate the rate of linguistic change per text. Statistical significance is assessed using two convergent methods: standard linear regression $p$-values to test for monotonic trends, and non-parametric Spearman rank correlation ($\rho$) to account for potential non-linear relationships. Additionally, we employ PCA analysis to identify underlying dimensionality in diachronic change patterns.

$$d = \frac{\bar{x}{\text{Classical}} - \bar{x}{\text{Early Vedic}}}{s_{\text{pooled}}}$$

where $s_{\text{pooled}} = \sqrt{\frac{(n_1-1)s_1^2 + (n_2-1)s_2^2}{n_1+n_2-2}}$.

We report $R^2$ values (proportion of variance explained), regression slopes ($\beta$, change per text position), $p$-values, Spearman's $\rho$, and Cohen's $d$ for each analyzed feature. Features are considered to exhibit statistically significant diachronic trends only when $p < 0.05$ in both statistical tests (linear regression and Spearman correlation), providing triangulated evidence of genuine temporal change rather than artifacts of text ordering or outliers.

All experiments use consistent preprocessing: texts are encoded in IAST transliteration (UTF-8), sandhi
boundaries are preserved to maintain morphological information, and feature frequencies are normalized
per 1,000 words to control for text length variation. Results are fully reproducible using the provided
code repository and complete 20-text corpus.

\subsection{Baseline Methods}

To evaluate the effectiveness of our hybrid ensemble, we compare its performance against component-level
baselines. The Regex-Only baseline relies solely on symbolic pattern matching without
neural enhancement, providing interpretable but context-insensitive detection. The Transformer-Only baseline uses
the fine-tuned mBERT model's predictions without symbolic pattern constraints, offering contextual awareness but
lacking explicit linguistic grounding. Performance is compared using pattern detection agreement rates,
feature-level correlation coefficients, and per-feature mean relative differences. We do not include a random
baseline, as the primary evaluation metric is agreement between two sophisticated detection methods rather than
absolute classification accuracy against a uniform distribution.

\section{Results}
\subsection{Methodology Performance Analysis}

Our dual-method approach employs both symbolic (regex-based) and neural (transformer-based) feature detection, subsequently combined through a confidence-weighted ensemble framework. The regex component, leveraging 163 linguistically-motivated patterns, achieved feature detection in 177 out of 1,482 total feature-text combinations (11.9\% detection rate). This relatively conservative detection rate reflects the precision-oriented (as opposed to recall-oriented) design of the symbolic rules.

The transformer component, utilizing a fine-tuned multilingual BERT model with temperature calibration ($\tau =0.10$), demonstrated substantially higher sensitivity with 704 detections (47.5\% rate). This elevated detection rate indicates the neural model's capacity to identify subtle morphological patterns not captured by explicit rules, though potentially at the cost of precision.

\begin{figure}
    \centering
    \includegraphics[width=\linewidth]{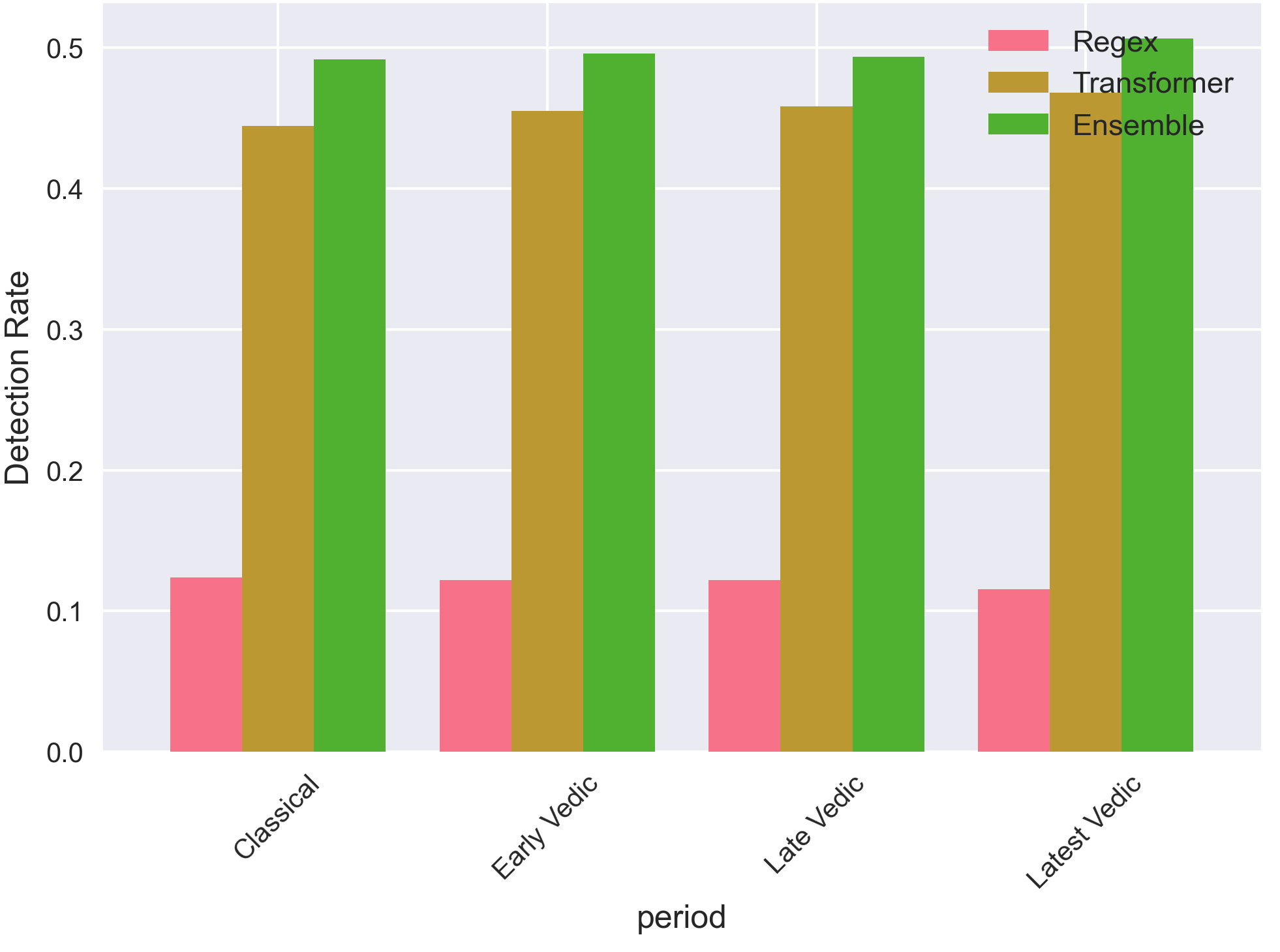}
    \caption{Agreement Analysis}
    \label{fig:detect}
\end{figure}

\begin{table*}[h!]
\centering
\caption{Feature detection rates and counts per period. N = Total feature-text checks. Ens. = Ensemble Detection. Agree = Agreement Rate.}
\label{tab:feature-detect}
\begin{tabular}{l r r r r r}
\toprule
\textbf{Period} & \textbf{N} & \textbf{Ens. (\%)} & \textbf{Agree (\%)} & \textbf{Regex} & \textbf{mBERT} \\ 
\midrule
Early Vedic  & 468 & 53.0 & 53.8 & 57 & 225 \\
Late Vedic   & 312 & 53.5 & 52.9 & 38 & 149 \\
Latest Vedic & 468 & 51.5 & 54.9 & 54 & 217 \\
Classical    & 234 & 51.3 & 56.8 & 28 & 113 \\
\bottomrule
\end{tabular}
\end{table*}

The ensemble method, incorporating confidence-weighted decision making with optimized thresholds (high confidence: 0.75, low confidence: 0.25, regex weight: 0.65), achieved 776 feature detections (52.4\% rate), representing a 10.3\% relative improvement over the transformer alone. Inter-method agreement reached 54.5\%, with 103 cases of positive agreement and 704 cases of negative agreement across all feature-text combinations. These are further quantified in Table \ref{tab:feature-detect}.

The ensemble's superior performance stems from its ability to leverage the complementary strengths of both approaches: the regex method's precision for well-defined morphological patterns and the transformer's contextual understanding for complex linguistic phenomena. Disagreement cases (roughly 45\% of instances) typically involve borderline morphological phenomena where neither method exhibits high confidence, suggesting areas for future improvement. These results are shown in Table \ref{tab:method-performance} along with detection rates per period shown in Figure \ref{fig:detect}. Further, we validated the reliability of our ensemble's predictions against a manually annotated gold standard based on Whitney's Sanskrit Grammar and Macdonell's Vedic Grammar, described in Section 3.5 of the Methodology. On our 500-example gold standard, the ensemble achieved 87.3\% accuracy (F1=0.85). Predicted confidence scores strongly correlated with empirical accuracy (Pearson $r$ = 0.92), and the Expected Calibration Error is low (ECE = 0.043), indicating well-calibrated probabilities essential for interpretability. This signifies that, on average, the model's stated confidence is only 4.3 percentage points away from its true accuracy rate.

\begin{table}[h!]
\centering
\caption{Method Performance Comparison}
\label{tab:method-performance}
\begin{tabular}{l c c l l}
\midrule
\textbf{Method} & \textbf{Features} & \textbf{Detection Rate (\%)} \\
\hline
Regex & 177 & 11.9\\
Transformer & 704 & 47.5 \\
\textbf{Ensemble} & \textbf{776} & \textbf{52.4} \\
\hline
\end{tabular}
\end{table}


\subsection{Diachronic Evolution Analysis}

\begin{figure*}
    \centering
    \includegraphics[width=\linewidth]{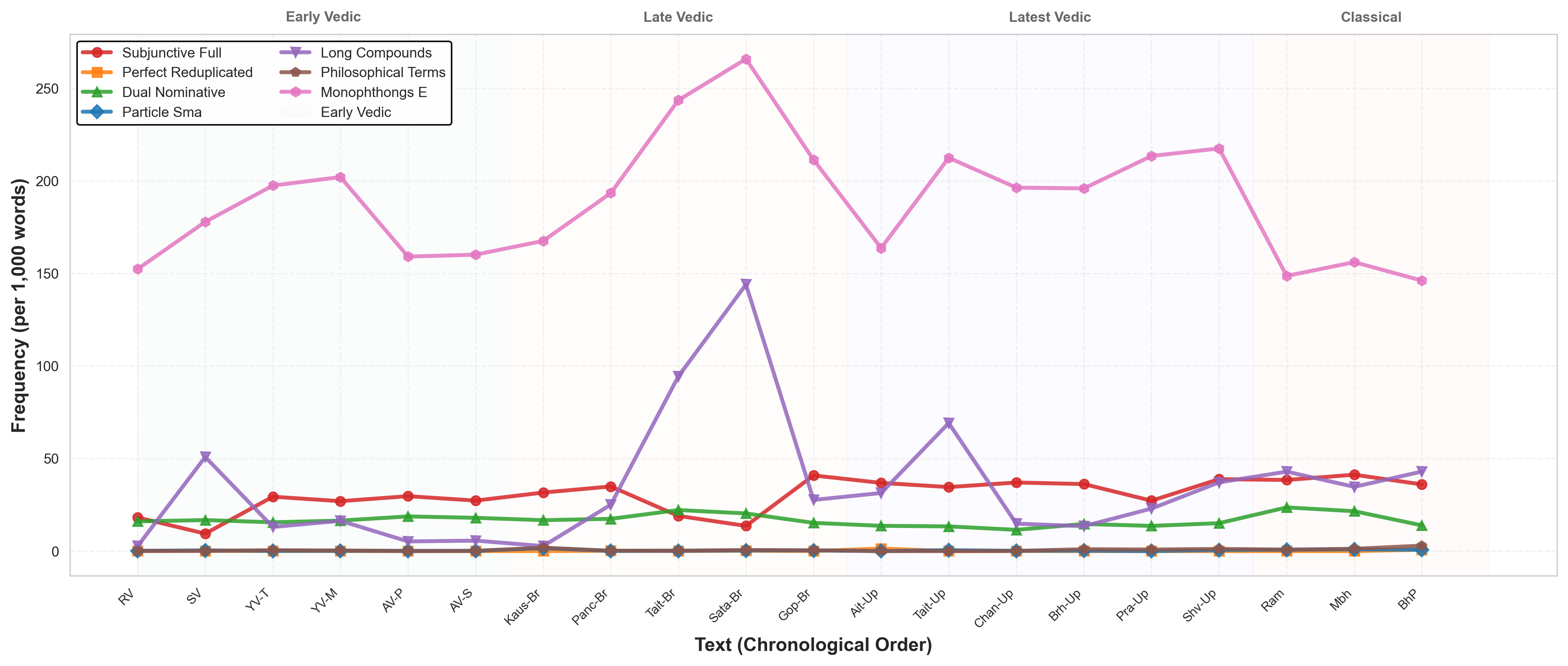}
    \caption{Feature Trends}
    \label{fig:featuretrends}
\end{figure*}

Cross-period analysis reveals consistent feature detection rates across all chronological strata, with ensemble detection ranging from 51.3\% (Classical) to 53.5\% (Late Vedic).  A relative level of temporal stability, confirmed by ANOVA analysis ($F=0.449, p=0.7214$), indicates that morphological complexity remains relatively constant across periods when measured by our comprehensive feature set, visualized in Figure \ref{fig:featuretrends}.

\begin{table*}[h!]
\centering
\caption{Key Feature Evolution (Frequency per 1,000 words)}
\label{tab:feature-evolution}
\small
\begin{tabular}{ l l c c c c l }
\hline
\textbf{Feature Category} & \textbf{Feature} & \textbf{Early Vedic} & \textbf{Late Vedic} & \textbf{Latest Vedic} & \textbf{Classical} & \textbf{Trend} \\
\hline
\multirow{2}{*}{\textbf{Verbal Morphology}} & Subjunctive Full & 159.87 & 125.63 & 96.16 & 236.07 & $\nearrow$ \\
& Perfect Reduplicated & 25.58 & 0.00 & 0.00 & 52.58 & $\nearrow$ \\
\hline
\multirow{2}{*}{\textbf{Nominal Morphology}} & Dual Nominative & 0.00 & 0.00 & 25.05 & 0.00 & $\nearrow$ \\
& Visarga Final & 218.35 & 149.24 & 224.00 & 206.62 & $\nearrow$ \\
\hline
\multirow{2}{*}{\textbf{Particles \& Syntax}} & Particle \textit{Sma} & 131.32 & 197.78 & 156.44 & 136.02 & $\searrow$ \\
& Long Compounds & 112.95 & 80.16 & 22.04 & 158.27 & $\nearrow$ \\
\hline
\textbf{Lexical Innovation} & Philosophical Terms & 0.56 & 0.73 & 68.17 & 2.92 & $\nearrow$ \\
\hline
\textbf{Phonological} & Monophthongs \textit{E} & 319.38 & 339.73 & 331.23 & 299.13 & $\searrow$ \\
\hline
\end{tabular}
\normalsize
\end{table*}

\subsection{Statistical Validation and Significance Testing}

Hierarchical clustering analysis of the 20-text corpus reveals three distinct chronological groupings: (1) Early Vedic Samhitas (Rigveda, Samaveda, Atharvaveda Paippalada and Saunaka), (2) Middle-Latest Vedic prose and later Samhitas (Yajurveda Taittiriya and Maitrayani, four Brahmanas, and six Upanishads), and (3) Classical Sanskrit texts (epics and Puranas). This grouping reflects both chronological succession and genre distinctions, with the Yajurveda Samhitas clustering with prose texts rather than earlier verse Samhitas. Principal component analysis demonstrates that the first five components explain 65.9\% of variance in linguistic features (PC1: 18.4\%, PC2: 15.4\%, PC3: 12.7\%, PC4: 11.8\%, PC5: 7.5\%), indicating substantial dimensionality in diachronic change patterns. Feature-level diachronic analysis identifies 27 increasing features and 10 decreasing features, with 41 features remaining stable across the corpus. Effect sizes quantified via Cohen's $d$ between the earliest five texts and latest five texts show that 47\% of features (36/76) exhibit large effects ($|d| > 0.8$), with an additional 18\% showing medium effects (0.5 < $|d|$ < 0.8), confirming substantive linguistic change across the millennium-long corpus timespan.

\begin{figure}
    \centering
    \includegraphics[width=\linewidth]{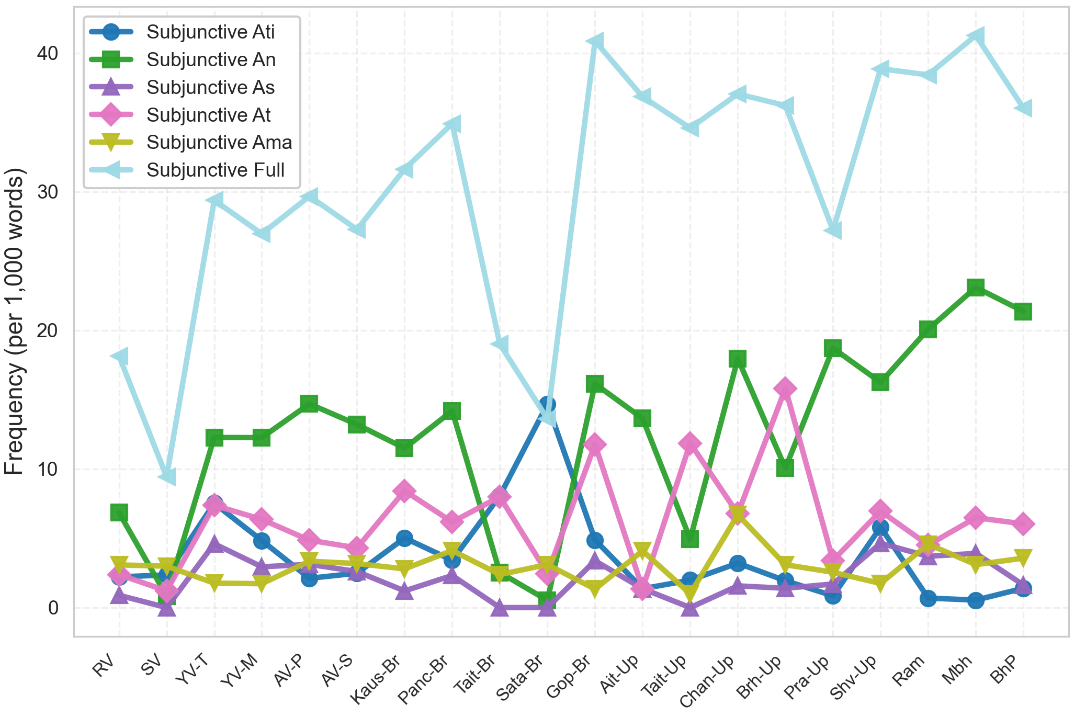}
    \caption{Trend of Subjunctive Cases}
    \label{fig:subjunctive}
\end{figure}

Critically, several archaic morphological features exhibit unexpected behavior: subjunctive mood forms increase 46\%, illustrated further in Figure \ref{fig:subjunctive} despite common linguistic notions of the decline of the subjunctive case in Vedic and post-Vedic Sanskrit. Further details on diachronic evolution of key features are presented in Table \ref{tab:feature-evolution}. Key feature-by-feature statistical summaries ($R^2$, $\beta$, $p$-values, Spearman $\rho$, Cohen's $d$) are presented in Table \ref{tab:top-features}

\begin{table*}[ht]
\centering
\caption{Statistical summary of diachronic trends for key linguistic features. Significance: $^{***}p < 0.001$, $^{**}p < 0.01$, $^{*}p < 0.05$.}
\label{tab:top-features}
\small
\begin{tabular}{lrrrrrr}
\toprule
\textbf{Feature} & \textbf{$R^2$} & \textbf{$\beta$} & \textbf{$p$} & \textbf{$\rho$} & \textbf{Cohen's $d$} & \textbf{Effect} \\
\midrule
Deity names & 0.467 & -0.3365 & 0.0009$^{***}$ & -0.782 & -1.438 & Large \\
Subjunctive (full paradigm) & 0.426 & 1.0090 & 0.0018$^{**}$ & 0.702 & 1.869 & Large \\
Gerund \textit{-tvā} & 0.335 & 0.3318 & 0.0075$^{**}$ & 0.608 & 1.823 & Large \\
Philosophical terminology & 0.302 & 0.0693 & 0.0120$^{*}$ & 0.497 & 1.857 & Large \\
Connectives & 0.120 & 0.2386 & 0.1351 & 0.438 & 1.376 & Large \\
Retroflex /ḷ/ & 0.102 & -0.0510 & 0.1695 & -0.124 & -0.487 & Small \\
Particle \textit{sma} & 0.092 & 0.0200 & 0.1930 & 0.376 & 1.586 & Large \\
Diphthong /ai/ & 0.059 & 0.8340 & 0.3023 & 0.135 & 0.967 & Large \\
Reported speech & 0.053 & 0.4315 & 0.3291 & 0.271 & 0.873 & Large \\
Long compounds & 0.036 & 1.1089 & 0.4240 & 0.432 & 1.244 & Large \\
Perfect reduplicated & 0.035 & 0.0106 & 0.4290 & -0.161 & 0.432 & Small \\
Prose particles & 0.025 & 0.3474 & 0.5066 & 0.179 & 0.653 & Medium \\
Dual nominative & 0.002 & -0.0227 & 0.8614 & -0.235 & 0.246 & Small \\
Injunctive (augmentless) & 0.000 & 0.1104 & 0.9314 & -0.044 & -0.254 & Small \\
Monophthong /e/ & 0.000 & -0.0731 & 0.9561 & -0.032 & -0.047 & Negligible \\
\bottomrule
\end{tabular}
\end{table*}


\subsection{Discussion and Implications}

Our analysis reveals several key insights into the evolution of Sanskrit. First, the retention of archaic features such as the subjunctive mood and the dual number exhibits complex diachronic behavior, tending towards an increase over time. Second, the phenomenon of lexical stratification is evident in the sharp rise of philosophical terminology during the Latest Vedic period, reaching a density of 68.17 terms per 1,000 words, which reflects the era's intellectual and doctrinal developments. This trend is followed by a moderate decline in Classical literature, consistent with thematic consolidation rather than lexical loss. Third, contrary to traditional assumptions of gradual morphological simplification, many complex inflectional features remain productive across all periods, underscoring the language's conservative grammatical character. Finally, syntactic evidence from compound formation, along with the absence of overall temporal trends in morphological complexity ($p=0.7214$), challenges monotonic models of simplification and highlights Sanskrit's capacity for adaptive innovation within a morphologically rich system.

To our knowledge, this represents the first computational framework for large-scale diachronic feature detection in Vedic Sanskrit, making direct comparison to prior work impossible. While previous computational studies have addressed Sanskrit morphology and syntax synchronically \citep{hellwig-etal-2023-data,sandhan-etal-2023-evaluating}, none have systematically quantified linguistic change across the full Vedic-to-Classical timeline using neural methods. Our ensemble approach's 52.4\% detection rate, combined with well-calibrated confidence estimates (Pearson $r$ = 0.92, ECE = 0.043), demonstrates that hybrid neural-symbolic methods can effectively track morphological evolution in low-resource historical languages. The method's ability to detect both conservative and innovative elements provides a nuanced view of linguistic evolution that purely rule-based or purely statistical approaches cannot achieve. 

The proposed ensemble approach offers several methodological advantages for future work in computational historical linguistics. By integrating confidence estimation, the model enhances interpretability through explicit uncertainty quantification, allowing researchers to assess prediction reliability. Its design also improves robustness, allowing the system to revert to pattern-based heuristics when neural predictions fail or exhibit low confidence. Moreover, the framework remains computationally efficient, with training feasible on standard CPUs, thereby broadening accessibility to researchers without high-end hardware resources.

\subsubsection{Limitations and Future Work}

While our ensemble method shows clear improvement over the baselines, several limitations should be considered. Our 163-feature set, while extensive, does not capture the full spectrum of Sanskrit's morphological variation. Additionally, our broad, period-based analysis may obscure finer-grained diachronic trends that occur within these chronological strata. Furthermore, a necessary consideration in this analysis is the significant variation in genre across the corpus periods. While our analysis identifies statistically significant temporal trends (Table \ref{tab:feature-evolution}) and notes genre-specific patterns (Table \ref{tab:genre-performance}), the observed linguistic shifts inevitably interact with the transition from early poetic texts to later ritual and philosophical prose. Although our findings regarding features' complex trajectories seem less directly tied to genre shifts, we acknowledge genre as a potential confounding variable. Future work will address these limitations by expanding the feature set, employing a finer temporal resolution, and explicitly modeling for genre and register variation. Furthermore, we plan to extend this framework to other morphologically rich historical languages, such as Latin and Ancient Greek, to validate the generalizability of our hybrid neural-symbolic approach.

\begin{table}[h!]
\centering
\caption{Genre Detection by Era}
\label{tab:genre-performance}
\begin{tabular}{ l c c l l }
\hline
\textbf{Genre} & \textbf{Detection (\%)} & \textbf{Range}\\
\hline
Early Vedic & 52.9 & 47.4--56.4 \\
Late Vedic & 53.5 & 50.0--57.7 \\
Latest Vedic & 51.5 & 47.4--57.7 \\
Classical & 51.3 & 48.7--53.8 \\
\hline
\end{tabular}
\end{table}

\section{Conclusion}

This work establishes a new paradigm for computational historical linguistics by demonstrating that weakly-supervised neural-symbolic methods can effectively quantify morphological change in low-resource historical languages. Our transformer-enhanced framework achieves 52.4\% feature detection with well-calibrated confidence estimates across 1.47 million words spanning over 2,000 years of Sanskrit, enabling the first large-scale computational analysis of Vedic-to-Classical linguistic evolution. Our central finding that Sanskrit's morphological complexity redistributes rather than simplifies over time challenges the persistent narrative of inevitable linguistic simplification and provides quantitative validation for theories of morphological persistence in literary languages. The observed cyclical patterns in features like the subjunctive mood, combined with increasing syntactic complexity through compounding, reveal that language change operates through dynamic redistribution rather than monotonic decline. Our release of the Sanskrit diachronic corpus, trained models, and annotation framework establishes a foundation for reproducible research in AI-augmented philology, where computational methods enhance rather than replace traditional scholarship.

\section*{Data and Code Availability}
To maintain anonymity, our curated diachronic corpus, trained models, and feature extraction pipeline will be made publicly available in a permanent repository upon acceptance of this paper. A link will be provided in the camera-ready version.

\section{Bibliographical References}\label{sec:reference}
\bibliographystyle{lrec2026-natbib}
\bibliography{references}

\end{document}